%% file: main.tex
\documentclass{article} 

\usepackage{main}  
\usepackage{times}  
\usepackage{helvet} 
\usepackage{courier}  
\usepackage[hyphens]{url}  
\usepackage{graphicx} 
\usepackage{hyperref}       
\usepackage{booktabs}       
\usepackage{amsfonts}       
\usepackage{amsmath}
\usepackage{amssymb}
\usepackage{subfigure}
\usepackage{amsthm}
\usepackage{bm}
\usepackage{nicefrac}       
\usepackage{multirow}
\usepackage{color}
\usepackage{xspace}
\usepackage{caption}
\iclrfinalcopy

\usepackage{times}
\input{math_commands.tex}

\usepackage{hyperref}
\usepackage{url}
\newcommand*\samethanks[1][\value{footnote}]{\footnotemark[#1]}

\title{Auxiliary Learning for Deep Multi-task Learning}

\author{Yifan Liu\thanks{First two authors contributed equally to this work.},  \quad Bohan Zhuang\samethanks, \quad Chunhua Shen\thanks{Corresponding author. E-mail: chunhua.shen@adelaide.edu.au}, \quad Hao Chen, \quad Wei Yin  \\
The University of Adelaide,\
Australia\\
}

\usepackage{graphicx}  
\newcommand*{\eg}{e.g.\@\xspace}
\newcommand*{\ie}{i.e.\@\xspace}
\newcommand*{\etal}{et al.\@\xspace}

\def\mF{{\mathcal F}}
\def\mD{{\mathcal D}}
\def\mA{{\mathcal A}}
\def\mL{{\mathcal L}}

\def\btheta{\boldsymbol{\theta}}
\def\bh{\bm{h}}

\def\bx{\bm{x}}
\def\by{\bm{y}}

\begin{document}
\maketitle
\begin{abstract}
Multi-task learning (MTL) is an efficient solution to solve multiple tasks simultaneously in order to get better speed and performance than handling each single-task in turn. The most current methods can be categorized as either: 
(i) hard parameter sharing where a subset of the parameters is shared among tasks while other parameters are task-specific; or (ii) soft parameter sharing where all parameters are task-specific but they are jointly regularized.
Both methods suffer from limitations: the shared hidden layers of the former are difficult to optimize due to the competing objectives while the complexity of the latter grows linearly with the increasing number of tasks.

To mitigate those drawbacks, this paper proposes an
alternative, where we explicitly construct an auxiliary module to mimic the soft parameter sharing for assisting the optimization of the hard parameter sharing layers in the training phase. 
In particular, the auxiliary module takes the outputs of the shared hidden layers as inputs and is supervised by the auxiliary task loss. During training, the auxiliary module is jointly optimized with the MTL network, serving as a regularization by introducing an inductive bias to the shared layers.
In the testing phase, only the original MTL network is kept.
Thus our method avoids the limitation of both categories.
We evaluate the proposed auxiliary module on pixel-wise prediction tasks, including semantic segmentation, depth estimation, and surface normal prediction with different network structures. The extensive experiments over various settings verify the effectiveness of our methods.  
\end{abstract}

\input{introduction.tex}

\input{related_work.tex}
\input{method.tex}

\input{experiments.tex}

\input{conclusion.tex}

\bibliographystyle{iclr}

\bibliography{main}

\end{document}

%% file: math_commands.tex
\usepackage{amsmath,amsfonts,bm}

\def\eqref#1{equation~\ref{#1}}

\def\1{\bm{1}}

\def\mA{{\bm{A}}}

\def\mD{{\bm{D}}}

\def\mF{{\bm{F}}}

\def\mL{{\bm{L}}}

\def\mT{{\bm{T}}}

\DeclareMathAlphabet{\mathsfit}{\encodingdefault}{\sfdefault}{m}{sl}
\SetMathAlphabet{\mathsfit}{bold}{\encodingdefault}{\sfdefault}{bx}{n}



%% file: introduction.tex
\section{Introduction}

With the development of deep learning, convolutional neural networks have achieved great progress, especially in the field of computer vision \citep{SzegedyVISW16,li2018deep,He2016DeepRL,huang2017densely}. However, a practical system to understand a visual scene usually needs to perform diverse tasks in an efficient way, especially for the applications on mobile devices. 
The multi-task learning (MTL) system, aiming to solve different tasks simultaneously, becomes a popular topic in recent years. 
Another reason for this success is the intrinsic dependencies of the related tasks that lay in the data. This motivates the use of multiple tasks as an inductive bias in learning systems.
One straightforward solution to build a MTL system is the hard parameter sharing \citep{baxter1997bayesian,mousavian2016joint,Kokkinos2017UberNetTA}. The typical framework is shown in Fig. \ref{fig:overall} (a).
Specifically, the hidden layers are shared to learn the unified representations for all tasks, while output layers are task-specific. The final objective function is usually a linear combination of each task. 
Minimization of a weighted sum of empirical risk is only valid if diverse tasks are not competing, which is rarely the case. Conflicting loss functions may cause gradients update in different directions for the shared parameters, which makes the hard parameter sharing MTL difficult to be optimized properly. 

Another way to create a MTL system is the soft parameter sharing \citep{nie2018mutual,yang2016deep}. The typical framework is shown in Fig. \ref{fig:overall} (c), where each task has its own model with its own parameters. 
The typical way to share representations is to add direct connections across multiple tasks. The parameters for one single task can get extra information from other tasks and be regularized by the training of other tasks.
However, there is an apparent limitation that the network complexity will grow linearly with the increasing number of tasks, which is very computationally expensive.

The solution proposed in this paper aims to overcome the drawbacks of both categories. To avoid the explosive complexity with the increasing number of tasks, we still use shared hidden layers to extract features for various tasks. However, different from the previous hard parameter sharing framework, we still enable the explicit extra regulations among different tasks. The proposed learning framework is shown in Fig. \ref{fig:overall} (b). Specifically, in the training phase, we design auxiliary modules to mimic the effect of soft parameter sharing strategy. Each auxiliary module takes the multi-level features from the shared hidden layers as inputs and is supervised by the corresponding auxiliary task objective. Then the auxiliary modules and the original MTL framework are jointly optimized. In this way, the auxiliary module serves as a regularizer by introducing an inductive bias to balance the shared and task-specific representations in the shared hidden layers. 
Moreover, it can provide direct hierarchical gradients during back-propagation to assist the optimization of the shared parameters.
It is worth noting that the auxiliary modules are removed during the inference stage without introducing any extra computational burden to the MTL framework. To explore a better auxiliary architecture, we also utilize the reinforcement learning \citep{nekrasov2018fast} to search auxiliary modules automatically.  

We apply the proposed auxiliary learning strategy to different network structures commonly used in the scene understanding task to validate its effectiveness. The experiments are conducted on NYUD-V2~\citep{silberman2012indoor} and SUNRGBD~\citep{song2015sun}, where we find that the proposed approach can improve the performance of all the tasks in MTL system compared with hard parameter sharing baselines. The proposed method can achieve similar or even better performance compared with single task baselines, indicating that the knowledge from different tasks is helpful for training. Besides, our results are better than state-of-the-art MTL systems with a simpler structure during inference.

%% file: related_work.tex
\section{Related work}

Multi-task learning has achieved great success in various tasks in deep learning, ranging from  computer vision \citep{mousavian2016joint,NIPS2018_7406,xu2018pad,long2017learning} to natural language processing \citep{deng2018multi,sanh2018hierarchical}. There are two typical ways of building a MTL system, including hard parameter sharing \citep{baxter1997bayesian,mousavian2016joint,Kokkinos2017UberNetTA}  and soft parameter sharing \citep{nie2018mutual,yang2016deep} (see Fig. \ref{fig:overall}). 

The former one builds an efficient way to handle the MTL by sharing some layers to encode the common feature and design some task-specific layers for each task. 
For example, 
UberNet \citep{Kokkinos2017UberNetTA} designs a framework to extract features from different levels in the encoder to handle various tasks.
However, there is still a performance gap between the hard parameter sharing MTL and the single task baselines.
And there are several ways to improve the performance of the hard parameter sharing MTL to make it comparable to single tasks. On one hand, some literature focuses on introducing extra computational modules and making use of auxiliary tasks, \eg, attention-aware distillation module~\citep{xu2018pad}, multi-modal fusion block~\citep{NIPS2018_7406}, and task recursive modules~\citep{zhang2018joint}. Such methods benefit from the inductive bias from related tasks and can achieve state-of-the-art performance with the cost of efficiency. On the other hand, some literature \citep{kendall2018multi,chen2017gradnorm,sener2018multi,guo2018dynamic} pays attention to improve the training of MTL by adjusting the task weights, where they jointly optimize the task weights and the network parameters. Kendall \etal \citep{kendall2018multi} regard task weights as uncertainty, and derive a principled
loss function which can learn a relative weighting automatically from the data. Chen \etal \citep{chen2017gradnorm} and Guo \etal \citep{guo2018dynamic} balance the task weights using some
manually formulated rules considering the gradients or the difficulty of the tasks.
Sener \etal \citep{sener2018multi} explicitly cast multi-task learning as multi-objective optimization, with the overall objective of finding a Pareto optimal solution.
Complementary to these works, we target on tackling the optimization difficulty of the shared parameters in MTL. We introduce a simple yet effective auxiliary learning strategy to introduce inductive bias for regularizing the training of the shared hidden layers.

The later one uses individual paths for different tasks with message passing across tasks. Usually the distance between the parameters of the model is regularized in order to encourage the parameters to be similar \citep{duong-etal-2015-low, yang2016trace}.
To develop better mechanisms, Yang \etal \citep{yang2016deep} propose to 
use tensor factorisation to split the model parameters into shared and task-specific parameters for every layer. 
Nie \etal \citep{nie2018mutual} also build a mutual adaptation module to achieve better performance for both tasks.  
A representative soft sharing strategy is the ``cross-stich'' unit \citep{misra2016cross}, which can automatically learn the sharing structure from data rather than requiring user trial and error.
However, the computational complexity will grow linearly with the increasing number of tasks. To avoid this limitation, we instead propose to utilize the spirit of soft parameter sharing by leveraging knowledge from auxiliary tasks to assist optimizing hard parameter sharing MTL.

In addition, previous works \citep{wang2015training,zhao2017pyramid,nekrasov2018fast} also show that adding auxiliary losses into intermediate layers can accelerate the training process and may boost the performance.
However, the scales as well as the positions of the additional supervision are very heuristic and may have negative impact on the final performance. In contrast, our proposed auxiliary modules allow the hierarchical auxiliary supervision directly propagates back.

%% file: method.tex
\begin{figure}
  \centering
  \includegraphics[width=0.9\textwidth]{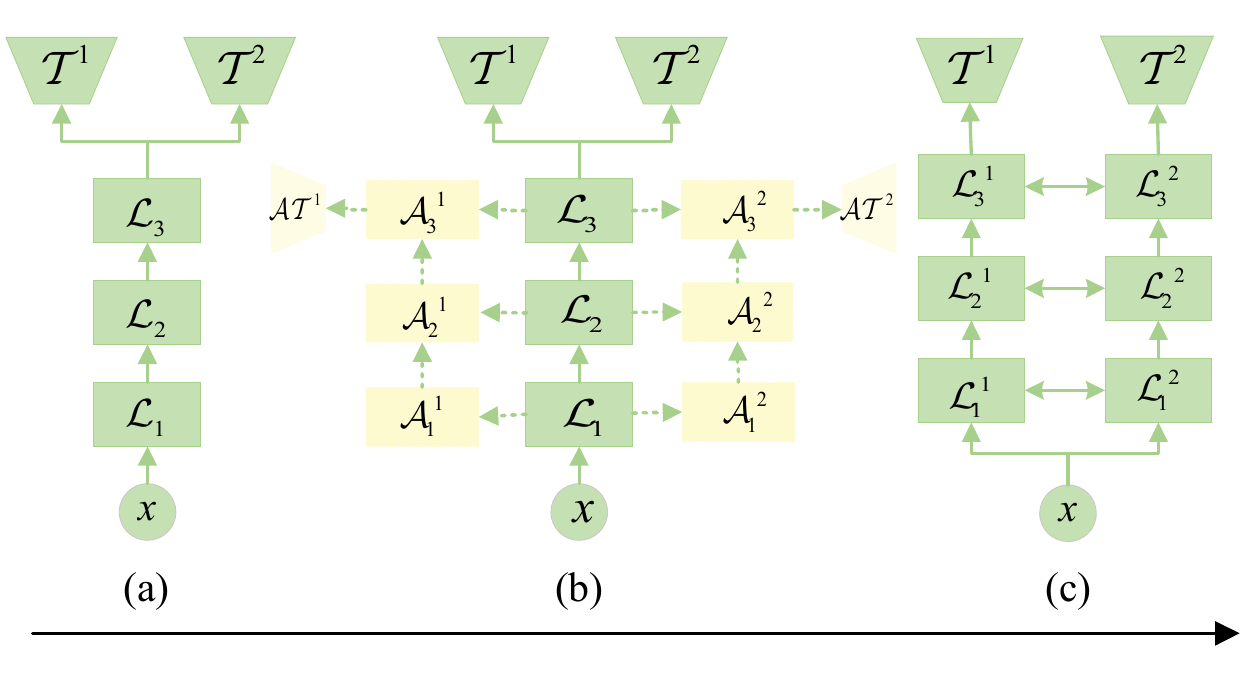}
  \caption{Comparison of two representative MTL methods and the proposed approach. We take two tasks as example. 
  (a) Hard parameter sharing: two tasks share the same layers to extract the features, and use task-specific layers to handle different tasks. (c) Soft parameter sharing: each task employs an independent network, but applies message passing between specific layers. Our proposed auxiliary learning method (b) lies somewhere between (a) and (c). We still use the hard parameter sharing framework for inference. But during training, the auxiliary modules mimic the soft parameter sharing to provide extra inductive bias to regularize the training for the shared layers.}
  \label{fig:overall}
\end{figure}

\section{Method}
We first give an overview of the hard parameter sharing MTL framework in Sec. \ref{sec:mtl}. Then we will describe the auxiliary module design in Sec. \ref{sec:auxi}.

\subsection{Problem definition}
\label{sec:mtl}

To train a multi-task learning system, we first need to get a dataset with the mapping from a single input space $\mathcal{X}$ to multiple labels $\{{\mathcal{Y}}^t\}_{t=1}^T$ , i.i.d. ${\{\bx_{i},\by_{i}^{1},\cdots,\by_{i}^{T}\}_{i=1}^N}$, where $T$ is the number of the tasks and $N$ is the number of the data samples. 
Specifically, the hard parameter sharing MTL network $\mathcal{F}$ consists of some shared layers $\mathcal{L}$ and $T$ groups of task-specific layers $\{\mathcal{T}^t\}^T_{t=1}$, whose parameters are defined as $\bm{\theta}^{\mL}$ and $\{\bm{\theta}^{\mT^{t}}\}_{t=1}^T$, respectively. Here we denote the loss function for the $t$-th task as $\ell^t(\cdot)$ to represent the difference between the output of the $t$-th task $f^{t}(\mathcal{X};\bm{\theta}^{\mL},\bm{\theta}^{\mT^{t}})$ and the corresponding ground truth label ${\mathcal{Y}}^t$. Therefore, we formulate the final objective function for a hard parameter sharing MTL network as follows:
\begin{equation}
\label{eq:mtl}
    \mathop {\min }\limits_{\{\bm{\theta}^{\mL},\bm{\theta}^{\mT^{1}},\cdots,\bm{\theta}^{\mT^{T}}\}}\sum_{t=1}^{T} \alpha^{t} \ell^t(f^{t}(\mathcal{X};\bm{\theta}^{\mL},\bm{\theta}^{\mT^{t}}),{\mathcal{Y}}^t),
\end{equation}
where $\alpha^{t}$ is a combination coefficient for the $t$-th task. In this work, we do not focus on adjusting the $\alpha^{t}$ to boost the performance, so we simply set $\alpha^{t}=1$ to all cases and reduce the objective function to $
    \mathop {\min }\limits_{\{\bm{\theta}^{\mL},\bm{\theta}^{\mT^{1}},\cdots,\bm{\theta}^{\mT^{T}}\}}\sum_{t=1}^{T} \ell^t(f^{t}(\mathcal{X};\bm{\theta}^{\mL},\bm{\theta}^{\mT^{t}}),{\mathcal{Y}}^t)$.

A major challenge in dealing with the optimization is the competing task objectives, which makes it difficult to optimize the shared parameters $\bm{\theta}^{\mL}$.
To tackle this problem, we will introduce an auxiliary learning strategy to assist the optimization of $\bm{\theta}^{\mL}$ in the next section.

\subsection{Auxiliary module}
\label{sec:auxi}

The proposed auxiliary learning framework during training is shown in Fig. \ref{fig:overall} (b). 
The key for our design is to mimic the effect of soft parameter sharing in Fig. \ref{fig:overall} (c) to  assist the training of the hard sharing parameters in  Fig. \ref{fig:overall} (a).
To achieve this, we explicitly utilize the weight sharing strategy to design the auxiliary modules. In particular, the $t$-th auxiliary module $\mathcal{A}^t$ takes the intermediate outputs of the shared hidden layers $\mathcal{L}$ as inputs and is supervised by the $t$-th auxiliary task loss. Then
the original MTL network $\mF$ as well as the auxiliary modules are optimized together.  
In the testing phase, the auxiliary modules $\{\mathcal{A}^t\}^T_{t=1}$ are removed without introducing extra computational burden to the original framework $\mathcal{F}$.

\noindent\textbf{Basic structure.}
The basic structure of the auxiliary module $\mA^t$ is made up of a sequential of adaptors and aggregators. 
In this paper, the auxiliary module takes $P$ outputs from shared layers $\mL$ as inputs. 
Let $\{ {L_1},...,{L_P}\}$ be the layer indexes where we add extra connection paths to the auxiliary module. For the output feature map ${\bf{O}}_{p}$ of the $L_p$-th layer, we apply a trainable adaptor $\mD^{p}(\cdot)$ to transfer the feature map into a task specific space and get the adapted feature $\mD^{p}({\bf{O}}_{p})$. We then use an operator $\odot$ to aggregate the adapted feature with the previous hidden representation $\bh_{p-1}$ to get $\bh_{p}$,  
\begin{equation}
 \bh_{p} =\bh_{p-1}\odot  \mD^{p}({\bf{O}}_{p}). 
\end{equation}
For the basic adaptor $\mD^{p}(\cdot)$, we choose a simple $1\times1$ convolutional layer followed with a batch normalization layer and the $\rm ReLU$ nonlinearity. The aggregate operators we employ are the concatenation or accumulation with respect to different architectures. 
The difference between the auxiliary training method and the soft parameter sharing is that the direction of the information flow is only from the main network to the auxiliary module. Therefore, the forward passes of the auxiliary module depend on the main network while the inference of main network does not rely on the auxiliary module. The auxiliary module $\mA^t$ can be treated as a regularization for inductive transfer, which mimics the scheme of the soft parameter sharing.

\noindent\textbf{Optimization.}
To jointly optimize the main network $\mathcal{F}$ and the auxiliary modules $\{\mathcal{A}^t\}_{t=1}^T$, we formulate the objective function as follows,
\begin{equation}
   \begin{aligned}
        &\mathop {\min }\limits_{\{\bm{\theta}^{\mL},\bm{\theta}^{\mT^{1}},\cdots,\bm{\theta}^{\mT^{T}},\bm{\theta}^{\mA^{1}},\cdots,\bm{\theta}^{\mA^{T}}\}}
        (\sum_{t=1}^{T} \ell^t(f^{t}(\mathcal{X};\bm{\theta}^{\mL},\bm{\theta}^{\mT^{t}}),\mathcal{Y}^{t}) \\
        & +\sum_{t=1}^{T} \ell^t_{aux}(\mathcal{A}^{t}(\bm{O}_{1},\cdots,\bm{O}_{L};\bm{\theta}^{\mL},\bm{\theta}^{\mA^{t}}),\mathcal{Y}^t)),
    \end{aligned}
\end{equation}
where $\btheta^{\mA^{t}}$ and $\ell^t_{aux}(\cdot)$ represent the adaptor parameters and the auxiliary loss for the $t$-th task, respectively.

From the equation, we can note that $\btheta^{\mL}$ is shared among $\mF$ and $\{\mathcal{A}^t\}_{t=1}^T$.
Following the chain rule, the gradients of the shared parameters will have an additional term comes from the regularization $\{{\ell}_{aux}\}_{t=1}^T$, which introduces an inductive bias from the auxiliary tasks. In this way, the hierarchical inductive bias can effectively balance the amount of shared and task-specific representations in the shared hidden layers to help generalize better.

\begin{figure}[tb]

  \centering
  \includegraphics[width=1.0\textwidth]{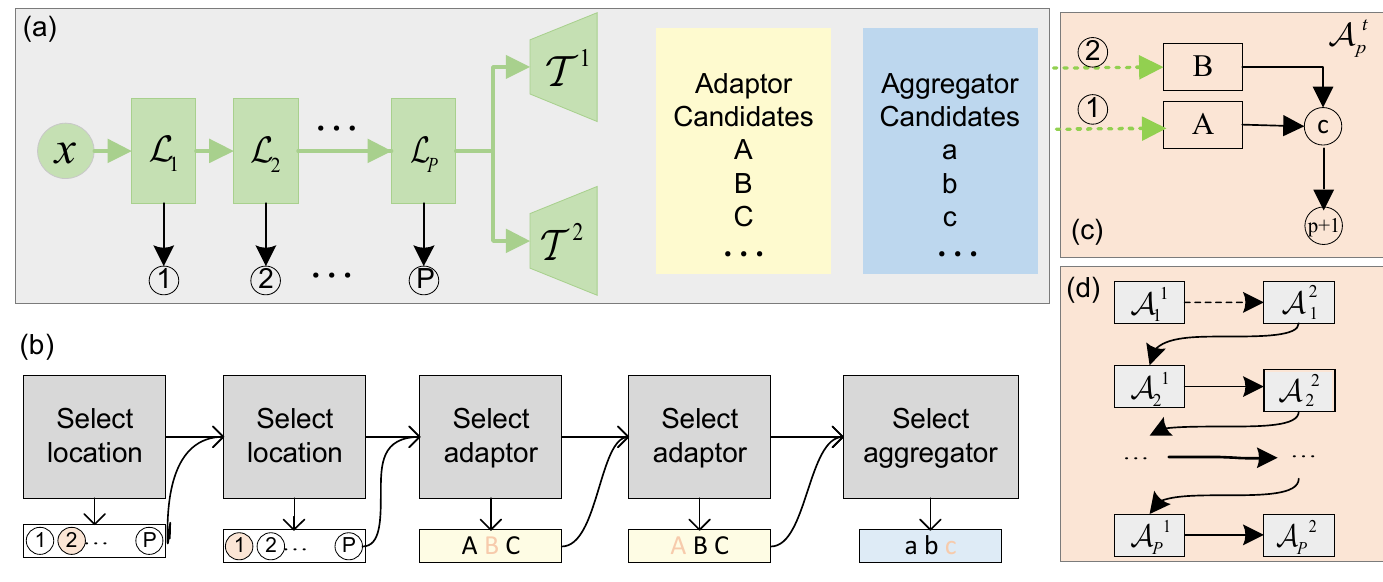}
  \caption{(a) Search space of NAS.  (b) Controller for generating the single auxiliary cell for the $p$-$th$ input of task $t$, $\mathcal{A}_p^t$. (c) An example of a single auxiliary cell generated by (b). (d) The order of generating the auxiliary cells recursively among different tasks.}
  \label{fig:nas}
\end{figure}

\noindent\textbf{Searching the auxiliary module.}
The basic structure of the auxiliary module is a feasible solution to assist the optimization of the shared parameters. However, the specific choices of the adaptors, aggregate operations and the connections of the auxiliary modules still need manual design.
Therefore, we further utilize neural architecture search (NAS) to explore the architectures of the auxiliary module in an automated way to remove hand-crafted heuristics.
The auxiliary module is used to help the training of off-the-shell MTL networks, therefore we include NAS only for searching the auxiliary module rather than the main network.
Moreover, the small searching space also makes the searching process very efficient.

The pipeline is shown in Fig.~\ref{fig:nas}. Fig.~\ref{fig:nas} (a) represents for the whole search space, including the input locations ($loc$), adaptor operations ($op_{ad}$) and the aggregate operations ($op_{ag}$). Two adaptors and an aggregate operation form an auxiliary cell. An LSTM-based controller is used to predict a sequence of operations and their locations. Each auxiliary cell is encoded by a sub-sequence of length $5$. As shown in Fig.~\ref{fig:nas}-(b), the controller first generates two indexes to choose nodes in $loc$ as the input nodes, then generates indexes from $op_{ad}$ for the adaptor operations. Finally, an index in $op_{ag}$ is sampled to aggregate two adapted features together to get an output node. The output of the auxiliary cell will be appended to $loc$ to serve as a new choice of input locations. Fig.~\ref{fig:nas}-(c) is an example of sampled auxiliary cell $\mA_{p}^t$ with the chosen output (colored in pink) in Fig.~\ref{fig:nas}-(b). The total output of controller is a sequence of length $5*P*T$, where $P$ is the number of auxiliary cells for each task, which equals to the number of auxiliary inputs.
Fig.~\ref{fig:nas}-(d) shows the order we search for the auxiliary cells. We generate auxiliary cells for each task following a hierarchical order by reserving the possibility of association between different tasks in the search space. For example, for the $p$-$th$ cell in the $t$-th task, it can not only choose the original $P$ intermediate feature maps as input but also choose the output from task $1$ to task $t-1$ in the first $p-1$ cells. In this way, the auxiliary module $\mA^t$ can learn a hierarchical relationships among all tasks. We show the search results of a sampled structure, which has the highest reward score on the validation set in Fig. \ref{fig:nas_de}.

Since the searching space for the auxiliary module $\mA^t$ is not as large as searching for a whole network, there is no need to make a trade-off by searching a sub-network then duplicate similar to~\cite{zhong2018practical, liu2018hierarchical, cai2018efficient, pham2018efficient}. 
To improve the searching accuracy, we jointly search for all the $T$ auxiliary modules $\{\mA^t\}_{t=1}^T$. 
Following the previous work of searching the structure for dense prediction problem \cite{nekrasov2018fast}, we include the following $6$ operators for adaptor candidates $op_{ad}$: separable $conv 3 \times 3$, $conv 1\times 1$, separable $conv 3 \times 3$ with dilation rate $3$, separable $conv 3 \times 3$ with dilation rate $6$, skip connections and deformable \cite{dai2017deformable} $conv 3 \times 3$. The aggregator candidates $op_{ag}$ include two operations: per-pixel summation and channel-wise concatenation of two inputs. Note that the basic structure is included in the designed search space.

After each sampling, we train the sampled network on the meta-train set and evaluate on the meta-val set following Nekrasov \etal~\cite{nekrasov2018fast} who have proposed an acceleration strategy for efficient NAS, especially with the small searching space. Therefore, in our work, it takes within one-gpu-day for searching $2000$ structures. The geometric mean of the evaluation metrics is employed as the reward. Gradients for the controller to maximize the expected reward are estimated with PPO\cite{schulman2017proximal}.

%% file: experiments.tex
\section{Experiments}

In this section, we explore the effect of our proposed auxiliary learning strategy on the hard parameter sharing multi-task learning. 
We first introduce the experiment setup in Sec. \ref{exp:multi_task_setup}, and then investigate on joint semantic segmentation and depth estimation from Sec.~\ref{exp:multi-task_strategies} to Sec.~\ref{exp:multi-task_comparison}. We further extend to three tasks by adding a head to predict surface normal in Sec.~\ref{exp:multi-task_sunrgbd}.
Following the previous works \cite{zhang2018joint,liu2019structured,Nekrasov2018RealTimeJS}, we employ the pixel accuracy (Pixel Acc.) and mean intersection over union (mIoU) to evaluate the segmentation task, mean absolute relative error (Rel Error) and root mean squared error (RMS Error) to evaluate the depth estimation and the mean error of angle (Mean Error) to evaluate the surface normal.

\subsection{Experiment setup} 
\label{exp:multi_task_setup}

\noindent\textbf{Datasets.} 
\emph{NYUD-v2} dataset \cite{silberman2012indoor} consists of $464$ different indoor scenes and has more than $100k$ raw data with depth maps, which are commonly used in the depth estimation task \cite{cao2017estimating,liu2016learning,fu2018deep}. $1,449$ images are officially selected to further annotate with segmentation labels, in which $795$ images are split for training and others are for testing. Following previous works \cite{Nekrasov2018RealTimeJS,zhang2018joint}, we also generate coarse semantic labels using a pre-trained segmentation network \cite{nekrasov2018light} for $4k$ randomly sampled raw data in official training scenes of \emph{NYUD-v2}, which only has the depth maps. We name this dataset as \emph{NYUD-v2-expansion.} 
We further conduct the experiments on \emph{SUNRGBD} \cite{song2015sun} dataset, which contains $10,355$ RGB-D images with semantic labels, of which $5,285$ for training and $5,050$ for testing. We follow the setting of previous work \cite{handa2016understanding} and perform the segmentation task with $13$ semantic classes.

\noindent\textbf{Network structures.} 
We employ the general hard parameter sharing structure by sharing the hidden layers (\ie, encoder) among all tasks while keeping several task-specific output layers (\ie, decoders).
We use the MobileNetV2 \cite{Sandler2018MobileNetV2IR} as the shared encoder by default. Three variants of the main networks are as follows: i) \emph{Baseline}: the baseline decoder is a two-layer task-specific classification module followed by a 8$\times$ bilinear upsampling layer. ii) \emph{Context}: We further add an ASPP module \cite{chen2018encoder} in the shared parameters to learn a stronger shared representation. iii) \emph{U-shape}: A U-Net structure decoder is added following the design of \cite{li2018deep}. And we use the ResNet-$50$ backbone to make a fair comparison with other state-of-the-art methods.
For MobileNetV2 \cite{Sandler2018MobileNetV2IR}, we take the output of each downsampling layer as the auxiliary input. For ResNet \cite{he2016deep}, the auxiliary module takes in the output of each residual block.

\noindent\textbf{Implementation details.} 
Common data augmentation is employed with random flip, random reshape (from $0.5$ to $2.1$) and random crop with the training size $385 \times 385$. The ground truth of depth should be normalized with the scale of random reshape. The batch size is set to $12$ for all experiments.
We verify our proposed method on the \emph{NYUD-v2} dataset in Sec.~\ref{exp:multi-task_strategies}. The models are trained for $30k$ iterations with the initial learning rate of $0.01$ and weight decay of $0.0001$. The learning rate is decayed by $(1-\frac{iter}{max-iter})^{0.9}$.  Apart from Sec.~\ref{exp:multi-task_strategies}, the models are pre-trained on \emph{NYUD-v2-expansion} for $40k$ iterations with the initial learning rate of $0.01$, and then fine-tuned on the \emph{NYUD-v2} with a fixed learning rate of $0.00001$ for $10k$ iterations. 

We further work on three tasks on the \emph{SUNRGBD} dataset, where the models are trained for $80k$ iterations with the initial learning rate of $0.01$ for both with and without auxiliary modules.

\subsection{Effect of different training strategies}    
\label{exp:multi-task_strategies}

\begin{figure}[htp]
  \centering
   \resizebox{0.8\linewidth}{!}
	{
		\begin{tabular}{c}
			\includegraphics{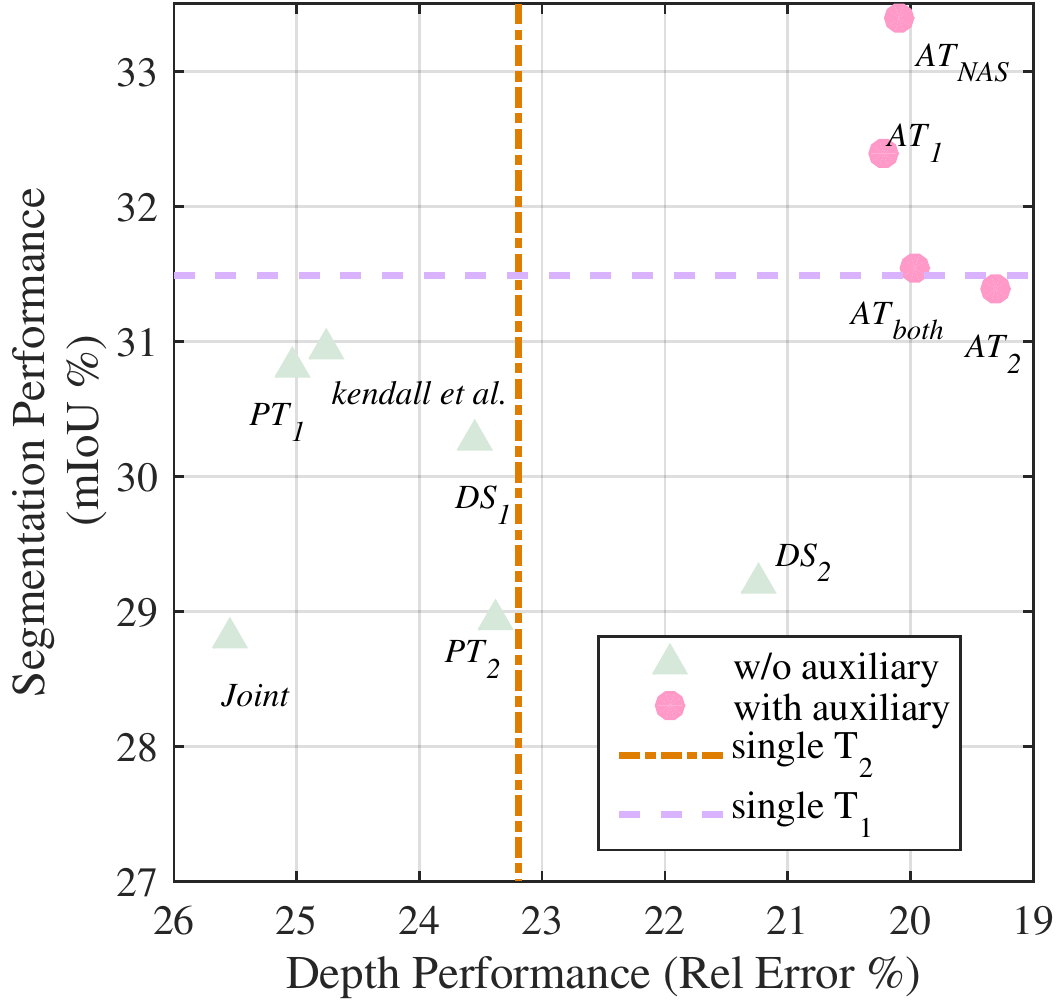}
		\end{tabular}
	} 
  \caption{\textbf{Performance of different training strategies.} We report the depth prediction and semantic segmentation results on the \emph{NYUD-v2}. Top-right is better. 
  We can see that adding auxiliary network can significantly boost the performance, even better than that of the single task.
  }
  \label{fig:baseline}
   \vspace{-1em}
\end{figure} 
 
In this section, we explore the effect of different training strategies on multi-task learning with semantic segmentation ($T_{1}$) and depth estimation ($T_{2}$). 
We include the following methods for study:

\begin{itemize}
    \itemsep -0.05cm 
    \item  \textbf{Single}: Solve each task independently (shown in Fig. \ref{fig:baseline} with the dotted line parallel to the coordinate axis and named \emph{Single}-$T_{i}$ in the following sections).
    \item \textbf{Joint}: Jointly train two tasks with the sum of two objectives.
    \item \textbf{Prior}: Initialize the MTL model from a pre-trained task ($T_{1}$ or $T_{2}$), and then jointly train two tasks (shown in Fig. \ref{fig:baseline} with $PT_{1}$ and $PT_{2}$, separately). The \emph{Prior} is widely used when we have a well-trained network, and want to initialize a multi-task case by adding a new sub-branch.
    \item \textbf{Kendall et al.} \cite{kendall2018multi}: Use the uncertainty weighting proposed by Kendall et al. \cite{kendall2018multi}.
    \item  \textbf{Deep supervision}: Train with additional losses supervised by $T_{1}$ or $T_{2}$ (shown in Fig. \ref{fig:baseline} with $DS_{1}$ and $DS_{2}$, separately). We add the additional losses at the end of each downsampling layer, and sum them all with scale of $0.1$. 
    \item  \textbf{Auxi}: Train with the auxiliary modules. In Fig.~\ref{fig:baseline}, 1) $AT_{1}$ or $AT_{2}$ represents for adding the basic auxiliary module with the supervision from $T_{1}$ or $T_{2}$, respectively. 2) $AT_{both}$ denotes for jointly optimizing the auxiliary modules for $T_{1}$ and $T_{2}$. 3) $AT_{NAS}$ represents we automatically search the auxiliary modules with two task supervisions. We name these three strategies as \emph{Auxi}-$T_i$, \emph{Auxi}-\emph{both}, and \emph{Auxi}-\emph{NAS} in the following sections.
\end{itemize}{}

To make a fair comparison, the strategies \emph{Prior}  ($AT_{1}$ or $AT_{2}$) and \emph{Auxi} for a single task ($AT_{2}$ or $AT_{1}$) are initialized with a pre-trained task ($T_{1}$ or $T_{2}$)\footnote{$AT_2$ represents that we add an auxiliary module for $T_2$ and initialize from $T_1$ to compare with $PT_1$.}. And the initial learning rate is divided by 10 for these cases.
All reported results are based on the \textit{Baseline} network.
From Fig.~\ref{fig:baseline}, we can see that directly jointly training the two tasks decreases the performance for both, which indicates the competing task objectives make the shared weights hard to optimize. 

The \emph{Joint} baseline can be boosted by adjusting the loss weighting following \emph{Kendall et al.} \cite{kendall2018multi}, but still performs worse than that of the \emph{Single} baseline.  Moreover, we also find that adding \emph{Deep supervision} has limited contribution to the performance.
The strategy \emph{Prior} introduces the prior knowledge of a specific task, thus improving the performance of the corresponding task compared with the \emph{Joint} baseline, but the performance of other tasks is still far below the corresponding \emph{Single} baselines. 

We can summarize three observations by comparing the proposed auxiliary learning method with other training strategies. First, with the auxiliary learning method, the performance of both tasks are significantly better than the \emph{Joint} baseline and other conventional training strategies, including learning task weights~\cite{kendall2018multi} and adding extra supervisions~\cite{zhao2017pyramid}. 
It can be attributed that the auxiliary modules can propagate effective regularization to the shared weights.
Second, adding auxiliary modules can make the performance even better than single task baselines, indicating that the auxiliary module transfers the inductive bias of the specific task, which assists the shared hidden layer to learn a better representation and generalize better.
Third, adding a single task-specific auxiliary module (\eg, $AT_{2}$) while being initialized from the prior knowledge of another task (\eg, $T_{1}$) can help improve the performance of auxiliary supervised task (\eg, $T_{2}$) compared with its corresponding \emph{Prior} baseline (\eg, $PT_{1}$).
Moreover, we also find that adding $AT_{both}$ can boost the performance for both tasks. However, for a specific task (\eg, $T_{1}$), it is slightly worse than that of only adding a single auxiliary module (\eg, $AT_{1}$).
To further replace $AT_{both}$ with $AT_{NAS}$, we get a significant $1.9\%$ improvement on mIoU for semantic segmentation task, which justifies that better auxiliary architectures are beneficial to assist the optimization of the shared parameters.

\begin{figure}[htb]
  \centering
  \resizebox{0.8\linewidth}{!}
	{
		\begin{tabular}{c}
			\includegraphics{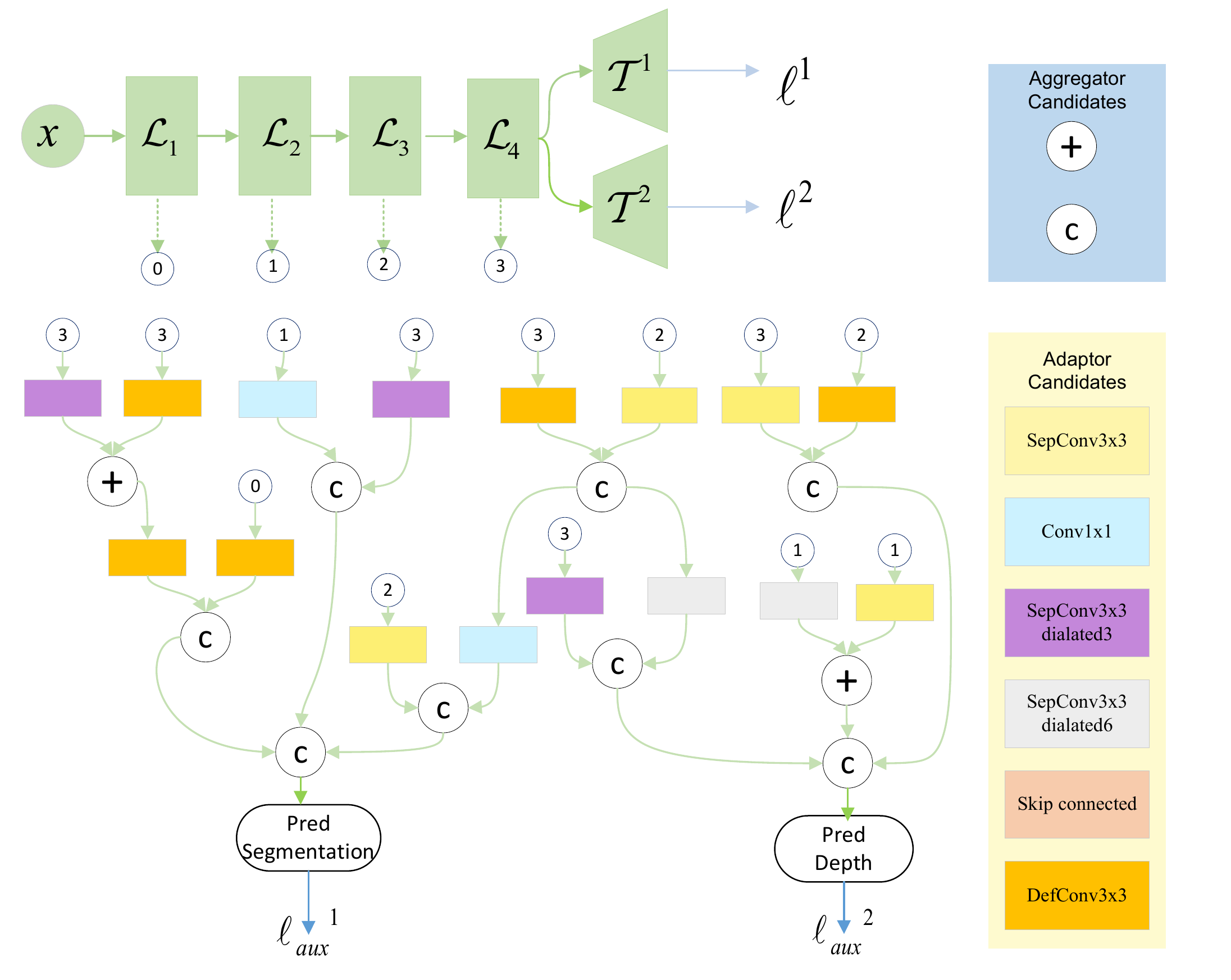}
		\end{tabular}
	}
  \caption{The auxiliary modules sampled by the reinforcement learning. We show the detailed structure for depth prediction and semantic segmentation.}
  \label{fig:nas_de}
\end{figure}

\noindent \textbf{Visualization of the sampled auxiliary module.}
We then visualize the auxiliary modules searched by \emph{Auxi}-\emph{NAS} in Fig. \ref{fig:nas_de}. We suppose the employment of NAS may offer interpretability for the auxiliary module. We can see the trend of each operator, for example, the number of ``skip connected'' operator decreases along the RL training, which indicates the ``skip connected'' is not helpful for the auxiliary module. Note that the search space for the auxiliary modules is small enough to make the searching process quite efficient. Moreover, we only target on improving the off-the-shelf hard parameter sharing MTL, therefore we do not simultaneously search the original framework. 

\begin{figure*}[ht]
  \centering
  \includegraphics[width=1.0\textwidth]{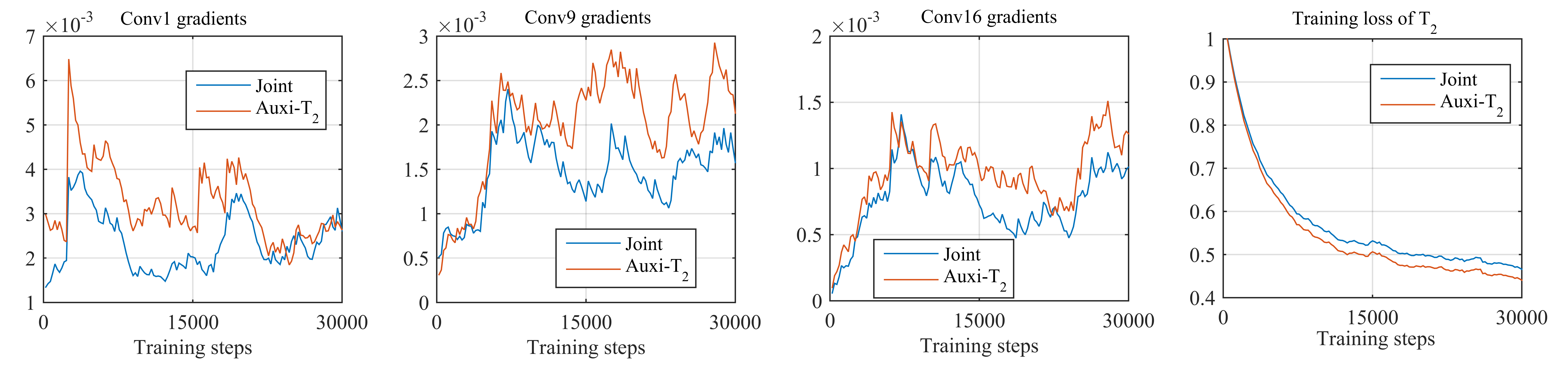}
  \caption{ \textbf{Training curves}. Base: jointly train two tasks. Auxi: adding a single auxiliary module supervised by the depth estimation. Here we show three samples of the average gradients for different layers (a-c) and the training loss curves for the depth estimation task. We can observe that the gradients w.r.t the shared parameters are enhanced.}
  \label{fig:curve}
\end{figure*}

\noindent \textbf{Visualization of training curves.}
To further analyze the benefit of training with the auxiliary module, we show the gradients in the shared layers and the loss curves of ${T_2}$ during  the training process for \emph{Joint} and  \emph{Auxi}-${T_2}$ in Fig. \ref{fig:curve}. 
We randomly sample some parameters from the shared convolutional layers to calculate the average gradients during the training process. The training loss curve of depth estimation is also included. From Fig. \ref{fig:curve}, we can observe adding auxiliary module provides extra gradients for the shared parameters, which may accelerate the optimization of the network \cite{nguyen2018extragradient}. Meanwhile, the training loss for $T_2$ is lower, which shows that the auxiliary learning may help the network convergence better.

\subsection{Effect of different main architectures}
\label{exp:multi-task_decoder}

In this section, we further explore the effectiveness of the proposed auxiliary learning strategy with different main network structures. We employ the structure,
which has the highest reward on the validation set during the searching process as our auxiliary module. 
All the structures are pre-trained on the \emph{NYUD-V2-expansion}, and then fine-tuned on the \emph{NYUD-V2}.  
We use \emph{Auxi}-\emph{NAS} for comparison 
and the results are shown in Tab. \ref{tab:decoder}.

\begin{table}[ht]
\footnotesize
\caption{Different network structures. $T_{1}$: Semantic segmentation.  $T_{2}$: Depth estimation.}
\centering
\scalebox{0.9}
{
\begin{tabular}{c|cc|cc|cc}
  \toprule
\multirow{2}{*}{Method} & \multicolumn{2}{c|}{Baseline}                           & \multicolumn{2}{c|}{Context}                            & \multicolumn{2}{c}{U-shape}                            \\
                        & \multicolumn{1}{c}{Rel Error} & \multicolumn{1}{c|}{mIoU} & \multicolumn{1}{c}{Rel Error} & \multicolumn{1}{c|}{mIoU} & \multicolumn{1}{c}{Rel Error} & \multicolumn{1}{c}{mIoU} \\
\hline

\emph{Single}-$T_1$  &  -  &      $34.6$&    -&$36.21$    &  -     &$38.05$                          \\
\emph{Single}-$T_2$ &   $18.7$&      -&   $17.13$ &-  &    $14.91$ &-                         \\
\emph{Joint}   &  $19.4$ &    $33.1$ &  $17.64$&$33.06$    & $15.12$&$37.38$     \\
Ours &  $\bf{18.6}$ &$\bf{34.8}$    &       $\bf{16.03}$&$\bf{37.13}$  &$\bf{14.64}$&$\bf{38.85}$\\
\hline
{\#Params (M)} & \multicolumn{2}{c|}{$3.31$}                           & \multicolumn{2}{c|}{$3.98$}                            & \multicolumn{2}{c}{$4.19$}                            \\
\bottomrule
\end{tabular}}
\label{tab:decoder}
\end{table}
From Tab. \ref{tab:decoder}, we summarize that as we increase the network's capacity, the performance of each task baseline boosts. It is a general way for one to get a higher performance while sacrificing speed. 
Moreover, for different main network architectures, the auxiliary modules can improve the performance for both tasks.
It justifies that the auxiliary module can provide hierarchical task-specific supervision for the shared backbone during training, which can assist the convergence of the hard parameter sharing MTL network. 
More importantly, even though the auxiliary module introduces extra parameters during training, we remove it during inference without introducing any complexity.

\subsection{Effect of different auxiliary architectures}
After testing on different variants of main networks, we then explore the influence of the architectures of the auxiliary module. We conduct experiments with the same main architecture \emph{Context}, and employ variants for the auxiliary module.
The results of various auxiliary modules with different structures are reported in Tab.~\ref{tab:capacity}. From the table, we can observe that increasing the complexity of the auxiliary module can further boost the performance. For example, by replacing the $1\times1$ convolution in the basic structure of adaptor with a large kernel of $3\times3$, we get performance gain on both two tasks by $0.18\%$ and $0.69\%$ respectively. For the \emph{Auxi}-\emph{both} method, the segmentation task has a relative $1.42\%$ improvement over the basic structure with $3\times3$ convolution. The reason can be attributed that a better auxiliary module architecture can provide more informative inductive bias for balancing the shared and task-specific representations in the shared hidden layers.  

\begin{table}[ht]

\caption{The performance w.r.t. different auxiliary architectures.}
	\centering
\begin{tabular}{l|c|c}
\toprule
Method&Rel Error&mIoU\\
\hline
\emph{Auxi}-\emph{both} with conv$1\times1$ &$16.12$&$35.02$\\
\emph{Auxi}-\emph{both} with conv$3\times3$ &$\bf{15.94}$&$35.71$\\
\emph{Auxi}-\emph{NAS}&$16.03$&$\bf{37.13}$\\
\bottomrule
\end{tabular}
\label{tab:capacity}
\vspace{-1em}
\end{table}

\begin{figure}[htp]
  \centering
  \includegraphics[width=0.9\textwidth]{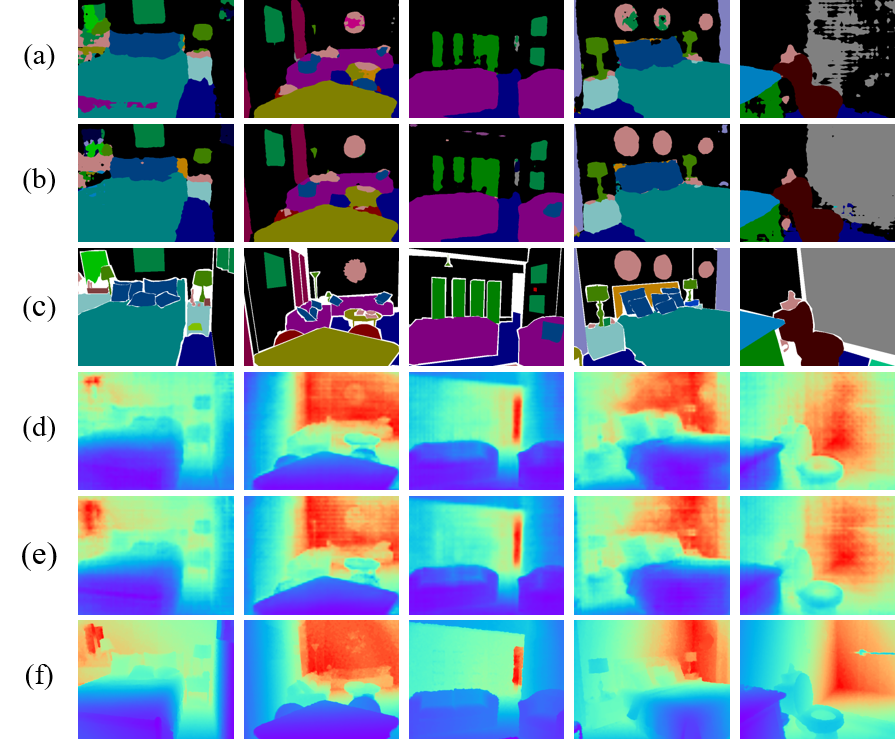}
  \caption{\textbf{Visualization results on NYUD-v2.} (a) and (d): Semantic segmentation and depth estimation results of ResNet50-joint. (b) and (e): Semantic segmentation and depth estimation results of ResNet50-Auxi-all. (c) and (f): Ground truth. With the help of auxiliary learning, the model can produce more consistent semantic labels for objects and more precious depth maps compared to the \emph{Joint} baseline.}
  \label{fig:vis_1}
\end{figure}

\subsection{Comparison with state-of-the-arts}    \label{exp:multi-task_comparison}
We further compare the proposed approach with other state-of-the-art methods designed for hard parameter sharing multi-task learning. We replace the backbone to ResNet-$50$, and employ task-specific decoders following \cite{yuan2018ocnet} for each task. The results are shown in Tab. \ref{tbl:benchmarks}. The auxiliary module can also boost the performance with a strong baseline such as ResNet-$50$. This justifies that the auxiliary learning can effectively solve the competing update for the shared parameters in the multi-task setting.
To compare with TRL-ResNet-$50$ \cite{zhang2018joint} which employs sophisticated decoders, we get better results over the mIoU, Rel Error and RMS Error with the proposed auxiliary learning even with the simplest decoders. Compared with the \emph{Joint} baseline which is jointly trained with two task losses, the proposed auxiliary learning strategy serves as an effective regularizer to help the convergence of the shared layers. We show some samples of visualization results in Fig. \ref{fig:vis_1}. We can empirically see that the semantic labels for objects are more consistent and the depth maps are more precise in our approach.

\begin{table}[htp!]

\caption{Results on the test set of \emph{NYUD-v2}. We show that our proposed auxiliary learning can improve the \emph{Joint} baseline to compare with other state-of-the-art methods designed for MTL. $T_1$: Semantic segmentation. $T_2$: Depth estimation. Err.: Error.}
\centering
\scalebox{0.98}
	{
		\begin{tabular}{l|c|cc}
         \toprule

			Method & mIoU  & Rel Err. & RMS Err. \\
			\hline
			Eigen and Fergus \cite{eigen2015predicting}&34.1&0.158&0.641\\
			Sem-CRF+\cite{mousavian2016joint}&39.2&0.200&0.816\\
			Real-Time \cite{Nekrasov2018RealTimeJS}&42.0&0.149&0.565\\
			TRL-ResNet50\cite{zhang2018joint}&46.4&0.144&0.501\\
			\hline
			ResNet-50-\emph{Single}-$T_1$&43.0&-&-\\
			ResNet-50-\emph{Single}-$T_2$&-&0.137&0.509\\
			ResNet-50-\emph{Joint}&42.9&0.143&0.563\\\hline
            ResNet-50-\emph{Auxi}-\emph{NAS}&\textbf{47.9}&\textbf{0.127}&\textbf{0.495}\\
	\bottomrule
		\end{tabular}}
	\label{tbl:benchmarks}
\end{table}

\subsection{Experiments on SUNRGBD} \label{exp:multi-task_sunrgbd}

\begin{figure}[htp!]
  \centering
  \includegraphics[width=0.9\textwidth]{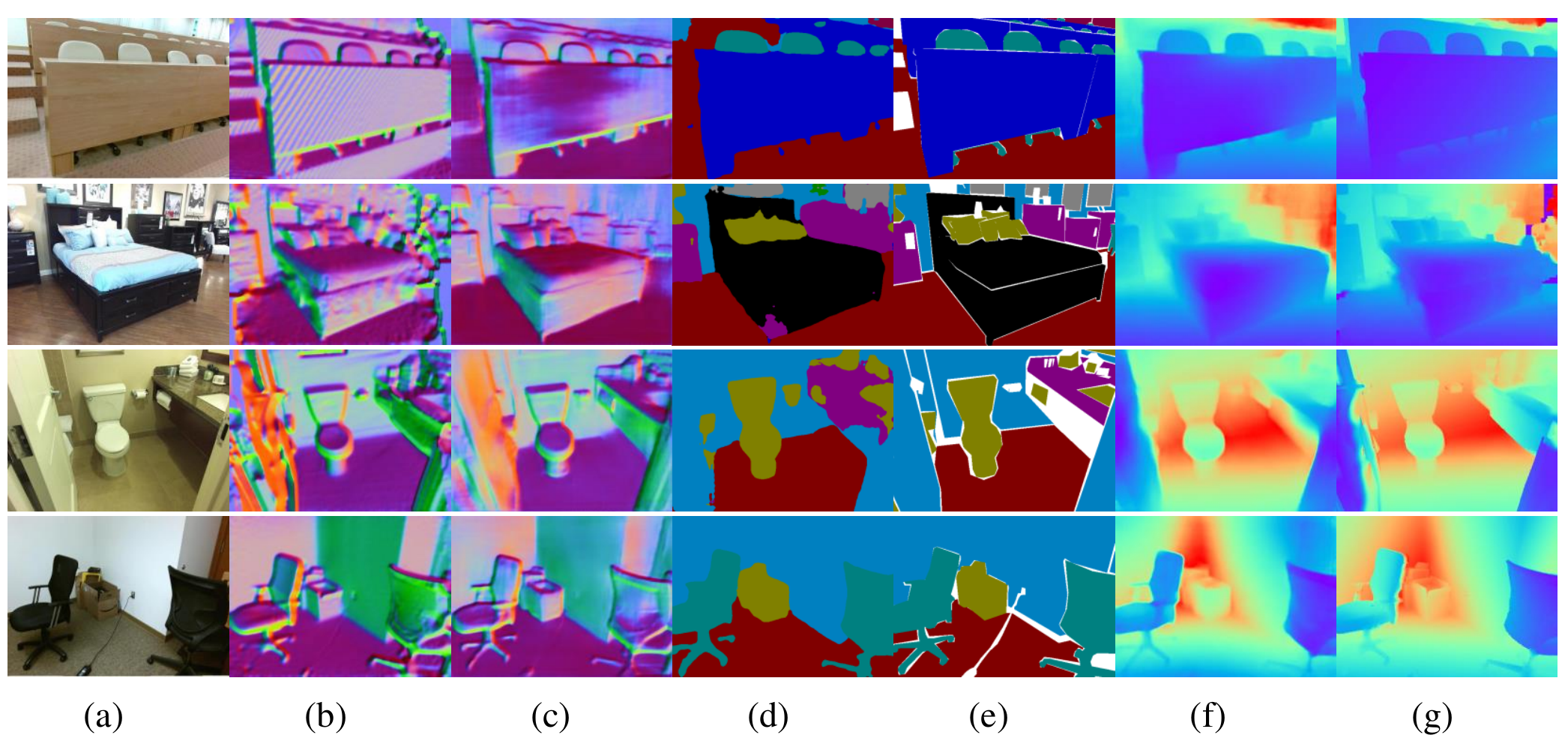}
  \caption{\textbf{Visualization results on SUNRGBD.} (a): Input image. (b): Predicted surface normal. (c): Ground truth surface normal. (d): Predicted semantic segmentation results. (e): Ground truth semantic segmentation results. (f): Predicted depth results. (g): Ground truth depth results.}
  \label{fig:vis_2}
\end{figure}

Finally, we conduct experiments on a larger dataset SUNRGBD with the proposed auxiliary learning method to verify its generalization ability. The experiments are based on a \emph{U-shape} MobileNetV$2$. And we further increase the number of tasks from two to three by adding a head performing the surface normal estimation. The results are reported in Tab. \ref{SUNRGBD}. We add the auxiliary module for each task separately or add them all together. We observe the auxiliary learning apparently boosts the performance for the specific single task. When combining them together, we can get an average improvement compared to the \emph{Joint} baseline. Besides, we can see that the depth prediction task and the surface normal estimation task are highly related, especially when we add the auxiliary module supervised by the surface normal loss, where the performance of the depth prediction is better than that of the \emph{Joint} baseline. It shows that the context information from the surface normal task plays an important role in the depth prediction task. We show visualization results on SUNRGBD in Figure~\ref{fig:vis_2}. The hard parameter sharing multi-task system with efficient MobileNetV2~\cite{Sandler2018MobileNetV2IR} backbone can generate multiple outputs including the surface normal, semantic segmentation maps and depth maps in one forward pass.

\begin{table}[htp]
\centering
\caption{Semantic segmentation, depth prediction and surface normal estimation results on the test set of SUNRGBD. \emph{Auxi}-$T_1$, \emph{Auxi}-$T_2$ and \emph{Auxi}-$T_3$ represent for adding a single auxiliary module with supervised loss from semantic segmentation, depth estimation and surface normal task, respectively. \emph{Auxi}-\emph{all} represents for adding them all together.}
\scalebox{0.9}
{
\begin{tabular}{c|cc|cc|cc}
\toprule
\multirow{2}{*}{Auxi type}& \multicolumn{2}{c|}{$T_1$: Segmentation}                           & \multicolumn{2}{c|}{$T_2$: Depth}                            & \multicolumn{2}{c}{$T_3$: Normal}                            \\
                        & \multicolumn{1}{c}{Pixel Acc.} & \multicolumn{1}{c|}{mIoU} & \multicolumn{2}{c|}{Rel Error} & \multicolumn{2}{c}{Mean Error ($^\circ$)} \\
\hline
 \emph{Joint} &  $80.7$& $53.7$&\multicolumn{2}{c|}{$22.8$}&\multicolumn{2}{c}{$28.7$}\\  
\hline

\emph{Auxi}-$T_1$&  {$\bf{82.9}$}  &{ $\bf{55.3}$}& \multicolumn{2}{c|}{  $21.7$}    &\multicolumn{2}{c}{$28.3$}                          \\
\emph{Auxi}-$T_2$&   $80.8$ &$54.0$ &\multicolumn{2}{c|}{$20.8$}&\multicolumn{2}{c}{$27.0$}                         \\
\emph{Auxi}-$T_3$ &$80.9$  &$54.4$&\multicolumn{2}{c|}{$\bf{20.3}$}&\multicolumn{2}{c}{$\bf{25.7}$}\\
\emph{Auxi}-\emph{all}&$81.3$  &      $54.9 $&    \multicolumn{2}{c|}{$20.5$}&\multicolumn{2}{c}{$26.1$}\\
\bottomrule
\end{tabular}}
\vspace{-1em}
\label{SUNRGBD}
\end{table}

%% file: conclusion.tex
\section{Conclusion}

In this paper, we have proposed an auxiliary learning strategy for the hard parameter sharing multi-task learning. Specifically, we have explicitly design auxiliary modules, which are supervised by the auxiliary task objectives. During training, the auxiliary modules are jointly optimised with the shared layers, which serves as the regularizer by introducing the hierarchical inductive bias to solve the competing gradient problem.
In the testing phase, we discard the auxiliary modules without increasing any computational complexity to the MTL framework. Moreover, we have utilized the neural architecture search to automatically explore the architectures of the auxiliary modules to avoid human heuristics.
We have empirically shown that such a training strategy can improve the optimization and generalization of the hard parameter sharing MTL system.